\pdfoutput=1

\documentclass[11pt]{article}

\usepackage{acl}

\usepackage{times}
\usepackage{latexsym}

\usepackage[T1]{fontenc}

\usepackage[utf8]{inputenc}

\usepackage{microtype}
\usepackage{hyperref}

\usepackage{xcolor}
\usepackage{tabularx}
\usepackage{multirow}
\usepackage{graphicx}
\usepackage{booktabs}

\usepackage{amsmath}
\usepackage{amsfonts}
\usepackage{multirow}
\usepackage{subfigure}
\usepackage{cleveref}

\usepackage{algorithm}
\usepackage{algpseudocode}

\title{Auto-MLM: Improved Contrastive Learning for Self-supervised Multi-lingual Knowledge Retrieval}

\author{
Wenshen Xu$^{1*}$, 
Mieradilijiang Maimaiti$^{1*}$, 
Yuanhang Zheng$^{2,3,4}$,
Xin Tang$^{1}$, 
and Ji Zhang$^{1\dagger}$
\\
$^1$Alibaba DAMO Academy \\
$^2$Department of Computer Science and Technology, Tsinghua University, Beijing, China \\
$^3$Institute for Artificial Intelligence, Tsinghua University, Beijing, China \\
$^4$Beijing National Research Center for Information Science and Technology \\
\tt{\{wenshen.xws,mieradilijiang.mea,eli.tx,zj122146\}@alibaba-inc.com}\\
\tt{zheng-yh19@mails.tsinghua.edu.cn}
}

\begin{document}
\maketitle

{
\let\thefootnote\relax\footnotetext{
$^*$ Equal contribution.}
\let\thefootnote\relax\footnotetext{
$^\dagger$ Corresponding author: Ji Zhang}
}

\begin{abstract}
Contrastive learning (CL) has become a ubiquitous approach for several natural language processing (NLP) downstream tasks, especially for question answering (QA). 
However, the major challenge,  how to efficiently train the knowledge retrieval model in an unsupervised manner, is still unresolved. 
Recently the commonly used methods are composed of CL and masked language model (MLM). 
Unexpectedly, MLM ignores the sentence-level training, and CL also neglects extraction of the internal info from the query. 
To optimize the CL hardly obtain internal information from the original query, we introduce a joint training method by combining CL and  Auto-MLM for self-supervised multi-lingual knowledge retrieval. 
First, we acquire the fixed dimensional sentence vector.
Then, mask some words among the original sentences with random strategy.
Finally, we generate a new token representation
for predicting the masked tokens.
Experimental results show that our proposed approach consistently outperforms all the previous SOTA methods on both AliExpress $\&$ LAZADA service corpus and openly available corpora in 8 languages.
\end{abstract}

\section{Introduction}
Among the natural language processing (NLP) downstream tasks, the question answering (QA) has became one of the ubiquitous branches.
More and more researchers have been focusing on the task of QA in both the academia and the industry. After the masked language model (MLM) has been proposed, pretrained language models like BERT \citep{bert} significantly improve the performances on various downstream tasks. In the research field of QA, pretrained language models also boost the models' performances \citep{DBLP:conf/emnlp/WangNMNX19,DBLP:conf/emnlp/HeZZCC20}.
Knowledge retrieval is also an important task in the research field of NLP. Especially in the research field of QA, knowledge retrieval is regarded as an essential subtask \citep{DBLP:conf/iclr/XiongXLTLBAO21,DBLP:conf/naacl/QuDLLRZDWW21,DBLP:conf/acl/YangWJJY20}.
To increase the speed, stability and accuracy of the QA systems, knowledge bases are often utilized in QA. Before providing answers for the users, the knowledge retrieval process should be performed on the knowledge bases.

\begin{figure}[!t]
\centering
\includegraphics[width=0.48\textwidth,height=4.1cm]{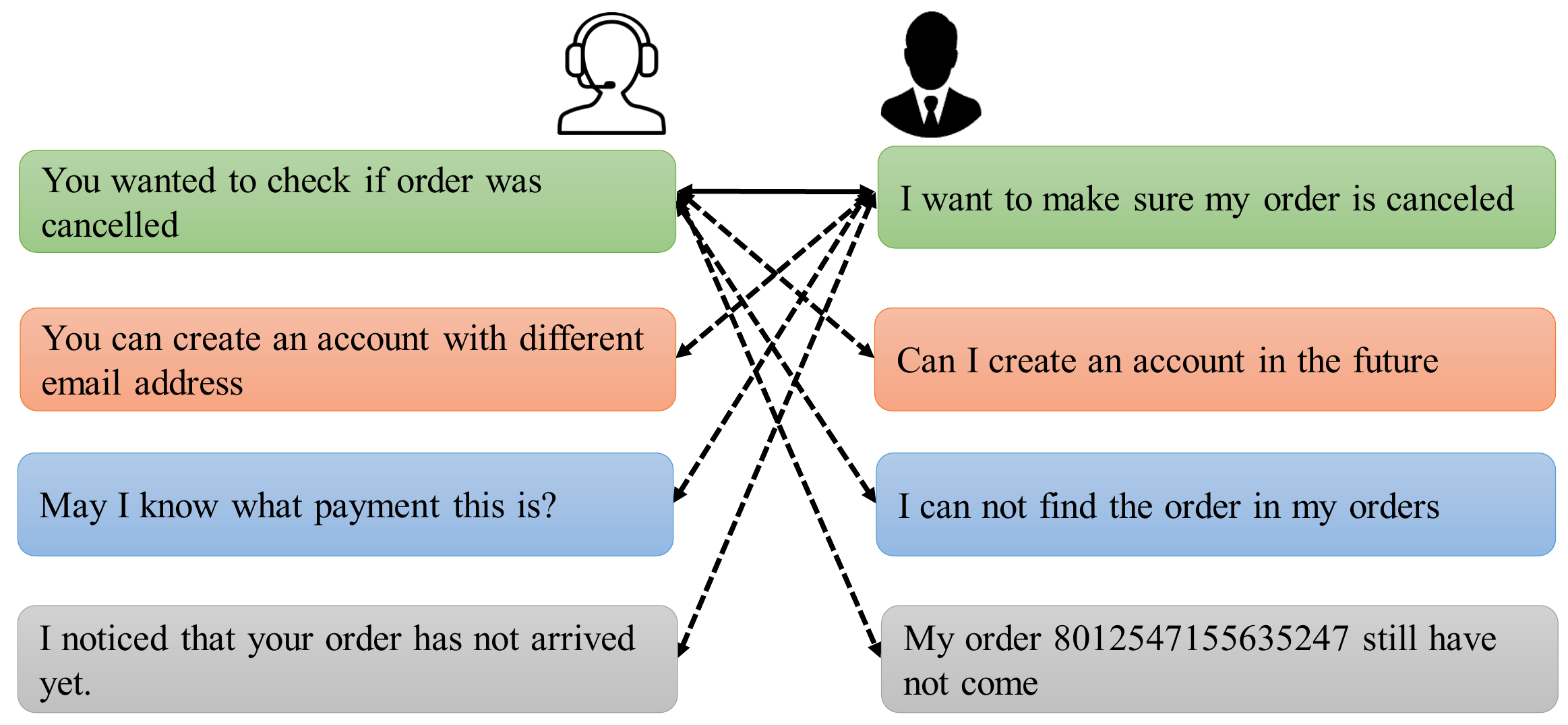}
\caption{Positive and negative samples for contrastive learning. {\em Left} side denotes agent query, and {\em Right} side represents user query. ``Solid line'' and ``dashed line'' denote positive and negative samples, respectively.} 
\label{fig_ctr_spl}
\end{figure}

In the dialog system, QA knowledge retrieval is an indispensable part of them. Therefore, we need to design a model which is qualified for extracting the semantic knowledge. However, in the multi-lingual QA scenario, it is always hard to train a model which has the better semantic extraction ability using limited supervised data. Generally, there are many methods such as conventional TF-IDF and neural network methods, but the mainly used tricks are combination of unsupervised corpus with pre-training methods.
Contrastive learning is a widely used training method of sentence semantic extraction in this field. However, contrastive learning is usually sensitive to the size of negative samples. Some work uses large negative sampling size \citep{Feng2020LanguageagnosticBS},
which brings some problems to hardware resources. Meanwhile, the context of human dialog systems usually contains a lot of noise, but constrastive learning is a sample-quality-sensitive training task and it will be unstable or hard to converge if there exists too much noise. Therefore, in the case of large data noise and limited hardware resources, it is necessary to explore other training signals to guide the model to learn better.

Currently, several pre-training methods \citep{Conneau2020UnsupervisedCR,Logeswaran2018AnEF,Gao2021SimCSESC,Yang2020UniversalSR} have been developed to utilize large-scale unsupervised corpus. Among these methods, \citet{Conneau2020UnsupervisedCR} and \citet{Yang2020UniversalSR} utilize MLM objectives, while the methodsd proposed by \citet{Logeswaran2018AnEF} and \citet{Gao2021SimCSESC} are based on contrastive learning. Though these methods can be used for knowledge retrieval in the multi-lingual QA scenario, they are mainly designed for general purposes, and none of them are specially designed for QA. The method proposed by \citet{Conneau2020UnsupervisedCR} is designed for fine-tuning on various downstream tasks, the method proposed by \citet{Logeswaran2018AnEF} is designed for text classification tasks, and the methods proposed by \citet{Gao2021SimCSESC} and \citet{Yang2020UniversalSR} are designed for retrieving semantically similar sentences. Thus, these methods may not perform well on QA tasks.

In this work we mainly discuss the task of self-supervised multi-lingual knowledge retrieval for QA via contrastive learning. As illustrated in Figure \ref{fig_ctr_spl}, both the negative and positive samples used in our approach are originated from user query and agent response.
The previous methods always ignore the internal information from the sentences. To deal with the above problem to some extent, we propose the \textsc{Auto-MLM} architecture. The model also can be regarded as a sentence-level MLM which is based on auto-encoder. 
Inspired by the awesome representation learning ability of MLM, we improve the model performance on sentence-level semantic extraction. 
We find that our proposed \textsc{Auto-MLM} approach also achieves significant results in non-context training corpus which cannot be used in CL, and also outperforms commonly used method MLM.
Precisely, our main idea highly keeps the consistency with the following steps.
First, we acquire the fixed dimensional sentence vector based on user query by taking advantages of {\em BERT} model. Then, mask some words among the original sentences with random strategy and encode the masked sentences.
Finally, we generate a new token representation by adding the achieved vectors of corresponding masked position and the original sentence vector obtained in the first step for predicting the masked tokens.

Our contributions are as follows:
\begin{itemize}
    \item We present the \textsc{Auto-MLM} architecture and fulfill the seamless combination of \textsc{Auto-Encoder} and \textsc{MLM}.  
    \item We also effectively improve the effectiveness of CL in QA knowledge retrieval task with Auto-MLM.
    \item We find the effect of \textsc{Auto-MLM} without CL is far better than the original MLM.
    \item Our jointly trained multilingual model obtains better improvements in all languages.
\end{itemize}

\begin{figure*}[!t]
\centering
\includegraphics[width=0.85\textwidth,height=6.5cm]{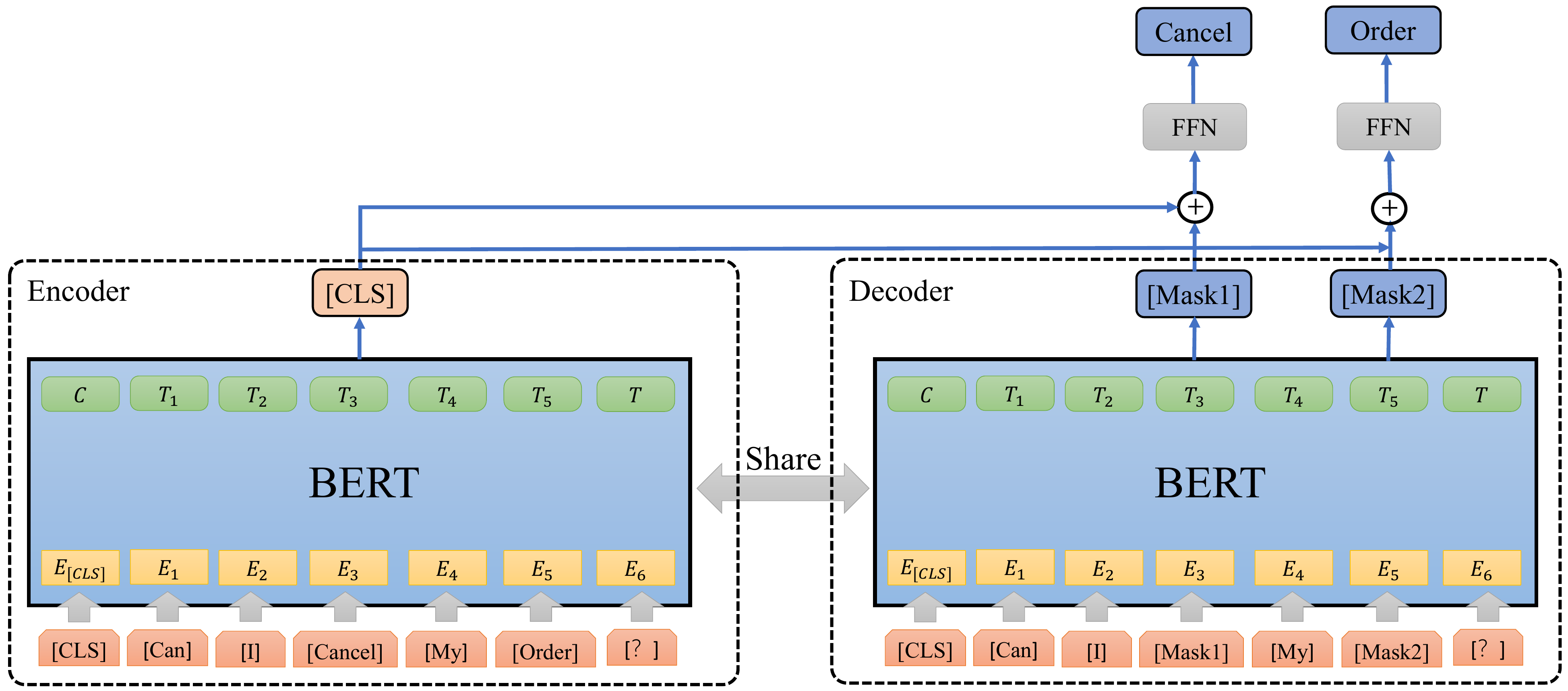}
\caption{The main architecture for our proposed Auto-MLM model. The {\em left} and {\em right} dashed rectangle can be seen as \textsc{Encoder} and \textsc{Decoder} of \textsc{Auto-Encoder}. The architecture of the decoder is the MLM.} 
\label{fig_auto_mlm}
\end{figure*}

\section{Background}

\subsection{Contrastive Learning}

Contrastive learning \citep{DBLP:conf/cvpr/HadsellCL06,DBLP:conf/icml/ChenK0H20} is a self-supervised representation learning method, which aims to pull similar samples together and push dissimilar samples apart. Formally, suppose the sample $x$ is similar to $x_1$ but dissimilar to $x_2$, and then contrastive learning aims to learn an encoder function $f$ such that
\begin{equation}
    sim(f(x),f(x_1)) >> sim(f(x),f(x_2)),
\end{equation}

\noindent where $sim(\cdot,\cdot)$ is a similarity function (e.g. cosine similarity).

When training models using contrastive learning methods, each batch contains a similar sample $x_1$ and $N-1$ dissimilar samples $x_2, x_3, \dots, x_N$. During training, we minimize
\begin{equation}
    \mathcal{L}=-\mathbb{E}_{x}\left[\log\frac{\exp(sim(f(x),f(x_1)))}{\sum_{i=1}^{N}\exp(sim(f(x),f(x_i)))} \right].
\end{equation}

Compared to generative self-supervised learning methods, a major advantage of contrastive learning is that it mainly focuses on high-level features of the samples, while generative methods may excessively focus on fine-grained features of the samples.

\subsection{Auto-Encoder}

Auto-encoder \citep{DBLP:journals/jmlr/VincentLLBM10,DBLP:conf/emnlp/SocherPHNM11} is a method for learning low-dimensional representation of high-dimensional data. An auto-encoder consists of an encoder and a decoder. For any high-dimensional sample $\mathbf{x}$, the encoder $enc(\cdot)$ encodes it into a low-dimensional representation $\mathbf{y}$, and the decoder $dec(\cdot)$ tries to recover $\mathbf{y}$ to a high-dimensional sample $\mathbf{x}'$. Formally,
\begin{align}
    \mathbf{x}'&=dec(\mathbf{y}) \nonumber \\
    &=dec(enc(\mathbf{x})).
\end{align}

When we train the auto-encoder, we expect that the restored sample $\mathbf{x}'$ is as close to the original sample $\mathbf{x}$ as possible. Therefore, during training, we minimize
\begin{align}
    \mathcal{L}=\Delta(\mathbf{x}',\mathbf{x}),
\end{align}

\noindent where $\Delta(\cdot,\cdot)$ is a distance function.

\subsection{Masked Language Model}

Masked language model (MLM) \citep{bert} is a training objective for learning representations of natural language sentences. When training a model using MLM objectives, we randomly mask some words in the sentence, and then the model is required to predict the masked words according to the remaining words in the sentence. Formally, for any sentence $\mathbf{x}$ in the training data, let $\mathbf{x}_m$ and $\mathbf{x}_o$ denote the masked and the observed part of $\mathbf{x}$. During the training process, we minimize
\begin{equation}
\label{eq:bert_loss}
    \mathcal{L}=-\mathbb{E}_{\mathbf{x}}\left[\log P(\mathbf{x}_m|\mathbf{x}_o;\mathbf{\theta})\right],
\end{equation}
\noindent where $\mathbf{\theta}$ is the model's parameters.

\begin{algorithm}[!t]
\caption{Improved CL with Auto-MLM} \label{alg:mixture_tl}
\begin{algorithmic}[1]
\Require multi-lingual mixed dataset $D_{mix}=\{\langle q_a,q_u \rangle \}_{s=1}^S$,  knowledge base $K_{mono}=\{\langle q^{(m)},c^{(m)}\rangle \}_{m=1}^M$, user query $Q_{user}=\{\langle \mathbf{q}^{(u)}\rangle \}_{u=1}^U$;

\Ensure retrieved $k$ similar question $Q^{\prime}$;

\For{N dialog-pair in $D_{mix}$} \Comment{batch-training}
    	\State achieve the loss $\mathcal{L}_{ctl}$ on (Eq.\ref{eq_ctl_lss})
    	\State obtain the loss $\mathcal{L}_{automlm}$ (Eq.\ref{eq_atml_lss})
    	\State jointly optimize $\mathcal{L}_{ctl} + \mathbf{\lambda} * \mathcal{L}_{automlm}$ (Eq.\ref{eq_joint_loss})
\EndFor
\State Get the current query $Q_{user}$ and  $K_{mono}$
\State Return retrieved $Q^{\prime}$ used $Q_{user}$ and $K_{mono}$

\end{algorithmic}
\end{algorithm}

\section{Methodology}
\label{lb_ses_method}

The core idea of our proposed method is a novel training approach to knowledge retrieval for QA task. As shown in Algorithm \ref{alg:mixture_tl}, we jointly train the model with CL and \textsc{Auto-MLM} objectives to improve the performance of knowledge retrieval model.
As depicted in Figure \ref{fig_auto_mlm}, our architecture consists of \textsc{Auto-MLM} and \textsc{CL}. Precisely, the \textsc{Auto-MLM} model is combined by \textsc{Auto-Encoder} and \textsc{MLM}.

\subsection{Contrastive Learning}
\label{lb_method_ctl}
We leverage the contrastive learning in dialog semantic extraction for QA,  and close the distance between queries in the same session and extend the distance between queries in different sessions. 
Since we assume that a session is usually to solve a problem, the intentions of all queries in the current session are closely related to each other.
For the sample $x$, we construct a sample $x_0$ from the same session and $N$ samples $x_1, x_2, x_3, \dots, x_N$ from different sessions.
In the training procedure of CL, we refer to the main idea of  LaBSE \citep{Feng2020LanguageagnosticBS}, and exploit the small trick ``in-batch" CL which views the other non-positive samples as the negative samples of current sentence.
    The exact details are shown as follow:
    \begin{enumerate}
        \item 
        sample one batch of samples $\{q_a,q_u\}_1,\{q_a,q_u\}_2, \dots, \{q_a,q_u\}_{N}$, where each pair $\{q_a,q_u\}$ belongs to the same session and different pairs belong to different sessions. $q_a$ and $q_u$ denote user input query and agent response, respectively.

        \item 
        We encode all the above pairs and achieve the corresponding sentence vectors using the output at the position of CLS. 
        Meanwhile, we concatenate all the $q_a$, $q_u$ vectors in each pairs based on their dimension separately, and obtain the vector $\mathbf{v}_{q_a}$ and $\mathbf{v}_{q_u}$ with the same dimension of batch size.

        \item We normalize the above mentioned vectors $\mathbf{v}_{q_a}$ and $\mathbf{v}_{q_u}$, then achieve the score by doing dot product.
        In this score matrix, the positive diagonal is the similarity of sentences in the same session, the others (other elements) are the similarity of negative examples.
        We optimize the dialog contrastive learning and minimize the cost function by leveraging margin loss \citep{Logeswaran2018AnEF}:

    \end{enumerate}
    
\begin{equation}
\begin{split}
    \mathcal{L}_{ctl}=1/N \sum_{i=2}^{N}\sum_{j \neq i} \big[ \max \big( 0, \underbrace{ cos\big(f(q_a^i), f(q_u^j)\big)}_{\small{\textrm{negative sample}}} \\ - \underbrace{cos\big(f(q_a^i), f(q_u^i)\big)}_{\small{\textrm{positive sample}}} - margin \big) \big],
    \label{eq_ctl_lss}
\end{split}
\end{equation}
where the subscription ``${ctl}$" denotes contrastive learning and ${f}$ represents the encoding function of our model.

\begin{table*}[!t]
\small
\begin{centering}
\begin{tabular}{l|lll|llllll}

\toprule
 \multirow{2}{*}{Corpora} &\multicolumn{3}{c}{Ali Express} &\multicolumn{6}{c}{LAZADA} \\\cline{2-10}
                          & Ar  & En & Zh & Ind & My & Ph & Th & Sg & Vi\\
\midrule
 Sessions &              $1.72$M & $1.09$M & $2.27$M & $2.21$M & $0.18$M & $0.23$M & $0.56$M & $63.30$K & $0.10$M   \\
 Chats     &             $9.15$M &  $28.81$M &  $25.61$M &  $11.22$M &  $4.55$M &  $7.39$M & $4.59$M & $1.67$M & $2.61$M    \\
Avg$_{Chat num}$  &      5.33 & 26.36 & 11.28 & 5.07 & 25.54 & 31.53 & 8.15 & 26.31 & 25.31   \\

\bottomrule

\end{tabular}
\par\end{centering}

\caption{Characteristics of our corpora. The ``Avg$_{Chat num}$" indicates the average chats number in each sessions.
}\label{tb:data_set}

\end{table*}

\subsection{Auto-MLM}
\label{lb_method_amlm}

MLM recently has become a ubiquitous representation learning paradigm, but MLM still suffers from modeling the sentence-level semantics. To deal with this issue, we propose the \textsc{Auto-MLM}. On the basis of the representation learning paradigm of MLM, auto-encoder is used to improve the sentence-level modeling ability of MLM. 
The proposed model \textsc{Auto-MLM} predicts the target words by exploiting the sentence vector and the output of masked position among the query.
    We use $\mathbf{x}_m$ and $\mathbf{x}_o$ to represent the masked and observable parts of the sentence $\mathbf{x}$. During the training procedure, we minimize the loss function:
\begin{equation}
    \mathcal{L}_{automlm}=-\mathbb{E}_{\mathbf{x}}\left[\sum_{i=1}^{n}\log P(\mathbf{x}_m^{(i)}|\mathbf{x}_o;f(\mathbf{x}), \mathbf{\theta})\right],
    \label{eq_atml_lss}
\end{equation}

\noindent where the ${\theta}$ is model parameter.

We aim to jointly train the above mentioned two objective functions:

\begin{equation}
    \mathcal{L} = \mathcal{L}_{ctl} + \mathbf{\lambda} * \mathcal{L}_{automlm}.
    \label{eq_joint_loss}
\end{equation}

\section{Experiment}
\subsection{Setup}
\subsubsection{Data Preparation}

All the languages used in our main experiment are Arabic (Ar), Chinese (Zh), English (En), Indonesia (Ind), Malaysia (My), Philippine (Ph), Thai (Th), Singapore (Sg), and Vietnamese (Vi). 
These languages are obtained from ALIBABA Ali Express $\&$ LAZADA.
Precisely, the first three languages Ar, Zh, and En are from Ali Express service corpus. While the other langauges Ind, Ml, Ph, Th, Sg and Vi are from ALIBABA LAZADA service corpus. 
Moreover we also exploit the openly available corpus AskUbuntu\footnote{\url{https://github.com/taolei87/askubuntu}} instead of using the service corpus.
Concretely, the dialog corpora of LAZADA and Ali Express are mixed as multi-lingual corpora.
The specifications of the corpora are listed in the Table \ref{tb:data_set}. Explicitly, these corpora come from the real dialog between the user and the agent in the e-commerce scenario. Usually after the user encounters a problem, they seek the help from an agent.
Since it is a spoken conversation, the data has various characteristics. For example, the conversation may be interrupted and paused, making it incomplete. 
Intuitively, the conversation is not strictly a question-and-answer dialog. 
The input may be incomplete or unclear, each role may be input continuously, and the current input may be a response to the above paragraphs far away. These problems bring difficulties in understanding semantic cohesion.

It is also important to explain our testset in this task.
For each language, our testset includes two files, one is the testset file, and the other is the Knowledge file.
The testset includes two columns: the first column is real user consultation, and the second column is the manually labeled corresponding knowledge. 
We search for the most similar question in Knowledge based on the user inquiries in the test to determine the knowledge of the question. we leverage the cosine similarity \citep{Mikolov2013EfficientEO} to calculate the similarity of sentences. 
Since the proposed model is used in the recall modules of QA, we mainly focus on whether the model could retrieve the correct knowledge.
In LAZADA and Ali Express QA scenarios, if the correct knowledge can be found in the set of model retrieved top-K candidates, we regard them as correct cases.
Moreover, 
if a character enters multiple sentences consecutively, we will use spaces for splicing and aggregation.
We use sentence-Piece\footnote{\url{https://github.com/google/sentencepiece}} \citep{Kudo2018SentencePieceAS} to segment the data. For the sentence-Piece dictionary, we continue to use the \textsc{XMLR} \citep{Conneau2020UnsupervisedCR} dictionary with a vocabulary size of 250k.
To consider the computation resources, we exploit the \textsc{TinyBert}\footnote{\url{https://github.com/huawei-noah/Pretrained-Language-Model/tree/master/TinyBERT}} \citep{Jiao2020TinyBERTDB} parameters to train all the models, e.g., hidden layers 4. The maximum length is set to 64 and the sentences exceeding this length will be truncated.
Additionally, we train our model on 2GPUs (V100) for 2 days, the batch size is 128, and the maximum number of training steps is 300k for model training.
Because our data has a lot of noise signals, we use a relatively small margin and set it to 0.1.

\subsubsection{Baselines}
\begin{itemize}

    \item \textsc{XLM-RoBERTa} (\textsc{XLMR}): \citet{Conneau2020UnsupervisedCR} use MLM self-supervised training and large-scale multilingual corpus to greatly improve the multilingual representation of the model. 
    
    \item \textsc{Contrastive Learning} (\textsc{CL}):  It is introduced in Quick-Thought \citep{Logeswaran2018AnEF}. 
    In context \textsc{CL}, the classic paradigm is that given a sentence and its context, 
    the classifier distinguishes the context from a set of candidate sentences.
    
    \item \textsc{SimCSE}: \citet{Gao2021SimCSESC} propose a self-predictive \textsc{CL} that takes an input sentence and predicts itself as the objective. 
    
    \item \textsc{Conditional MLM} (\textsc{CMLM}): \citet{Yang2020UniversalSR} present a method of modeling the relationship between adjacent sentences based on MLM.

\end{itemize}

We highly follow the main idea of these baseline methods with totally unsupervised way for multi-lingual knowledge retrieval task.
For fair comparison, we train all the baselines on the same corpora (e.g.,  Ali Express and LAZADA). Except for the \textsc{XLMR}$_{base}$, we train all the baselines with  \textsc{XLMR}$_{tiny}$ model.

\begin{table*}[!t]
\begin{centering}
\small
\begin{tabular}{ll|lll|llllll|l}
\toprule
\multirow{2}{*}{Methods} &  & \multicolumn{3}{c|}{Ali Express} &\multicolumn{6}{c|}{LAZADA}& \multirow{2}{*}{Avg.}\\\cline{3-11}
                                        &    & Ar  & En & Zh & Ind & My & Ph & Th & Sg & Vi & \\
\midrule
\multicolumn{2}{l|}{\textsc{XLMR}$_{tiny}$ \citep{Conneau2020UnsupervisedCR}} &74.6&66.5&89.7&51.3&64.4&51.6&50.8&63.3&48.5&62.3 \\
\multicolumn{2}{l|}{\textsc{XLMR}$_{base}$ \citep{Conneau2020UnsupervisedCR}} & 80.7&70.4&93.1&54.1&67.0&55.1&63.8&66.7&53.9&67.2 \\
\multicolumn{2}{l|}{\textsc{SimCSE} \citep{Gao2021SimCSESC}} &79.8&74.8&95.8&62.5&76.2&63.0&63.0&73.0&52.9&71.2 \\
\multicolumn{2}{l|}{\textsc{CMLM} \citep{Yang2020UniversalSR}} &72.8&70.9&84.6&53.8&74.3&63.4&55.1&71.7&53.5&66.7 \\
\multicolumn{2}{l|}{\textsc{CL} \citep{Logeswaran2018AnEF}}  &81.4&78.6&95.8&68.5&86.2&75.0&75.0&78.9&62.5&78.0 \\
\midrule
\multirow{2}{*}{Our work}& \textsc{Auto-MLM} & 74.2&74.3&94.6&61.8&77.8&65.0&63.8&75.0&58.7&71.7 \\\cline{2-11}
                         & \textsc{CL} + \textsc{Auto-MLM} & \textbf{81.7} & \textbf{80.6}& \textbf{97.0} & \textbf{72.5} & \textbf{87.6} & \textbf{76.7} & \textbf{79.3} & \textbf{83.1} & \textbf{68.0} & \textbf{80.7} \\
\bottomrule

\end{tabular}
\par\end{centering}

\caption{The comparison with accuracy score between baseline systems on \textsc{Top-30} queries from Ali Express and LAZADA corpus. The ``Avg." denotes the average score.
}\label{tb:top30}

\end{table*}

\begin{figure*}[htp]
\centering 

\subfigure[\textsc{\textsc{Top-20}}]{
\begin{minipage}{7.1cm}
\centering
\includegraphics[width=7cm,height=3.5cm]{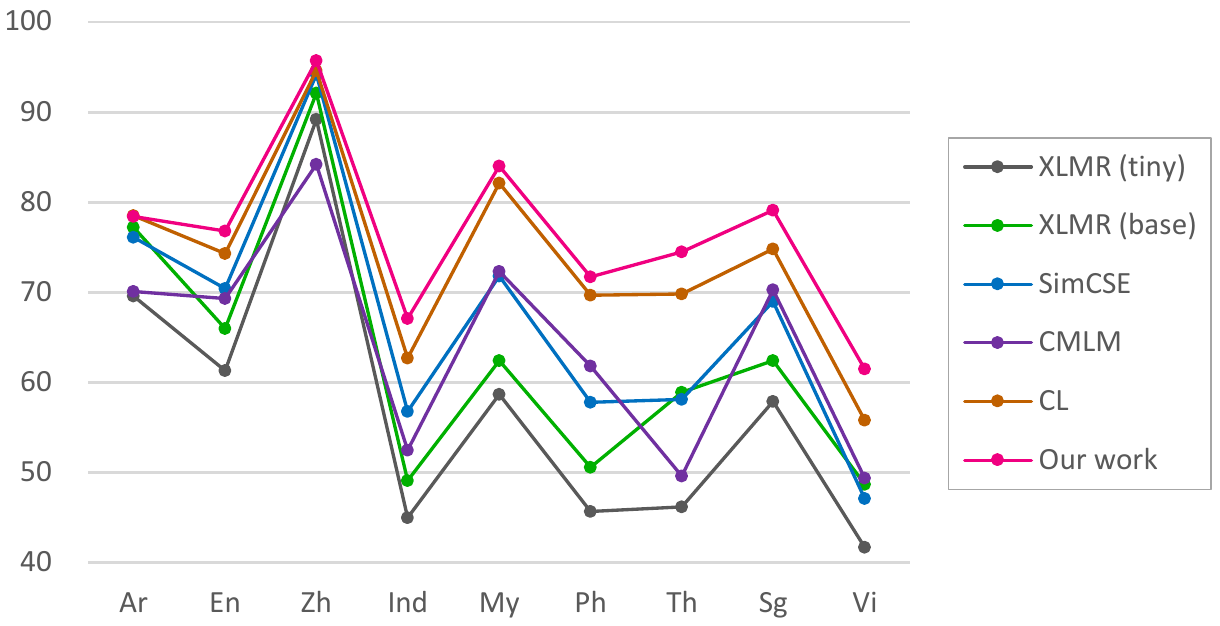}
\end{minipage}
}
\subfigure[\textsc{\textsc{Top-10}}]{
\begin{minipage}{7.1cm}
\centering
\includegraphics[width=7cm,height=3.5cm]{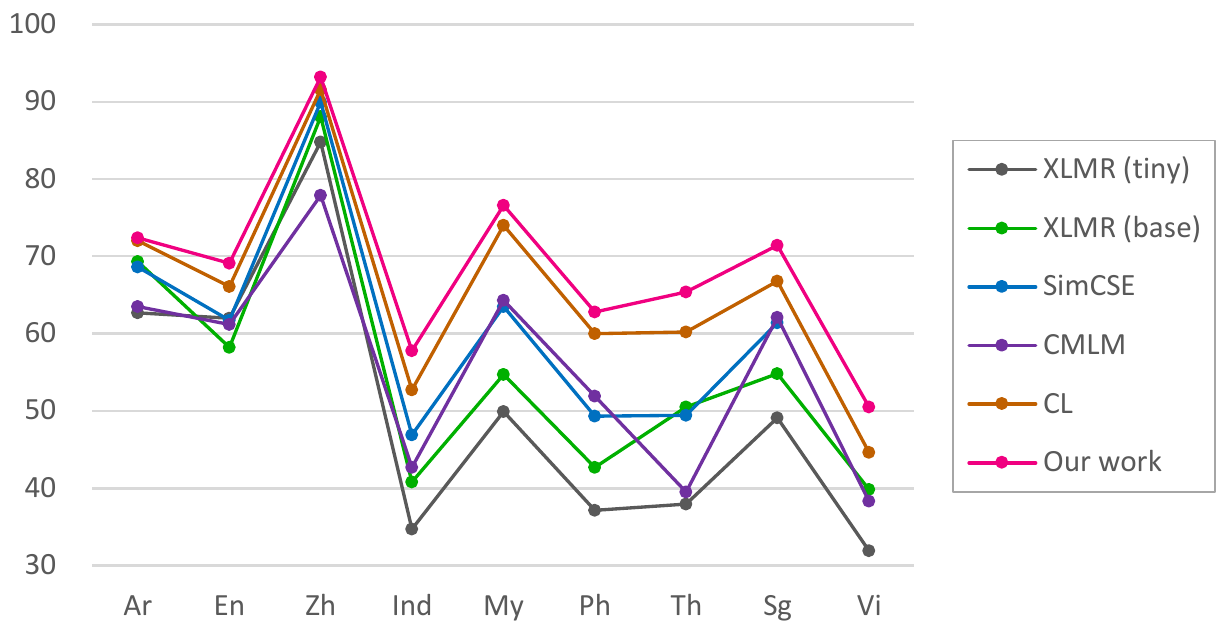}
\end{minipage}
}

\caption{The comparison with accuracy score between baseline systems on \textsc{Top-20} \& \textsc{Top-10} queries from Ali Express and LAZADA corpus.}

\label{lb_top_20_10}
\end{figure*}

\begin{table}[!t]
\small
\begin{centering}
\begin{tabular}{l|r|r}

\toprule
 \multirow{2}{*}{Methods} &\multicolumn{2}{c}{Askubuntu}\\\cline{2-3}
                          & p@1 & p@5\\
\midrule
\textsc{XLMR}$_{tiny}$ \citep{Conneau2020UnsupervisedCR} & 48.9 & 38.7\\
\textsc{SimCSE} \citep{Gao2021SimCSESC} & 51.6 & 38.0\\
\textsc{CMLM} \citep{Yang2020UniversalSR} & 52.7 & 40.2\\
\textsc{CL} \citep{Logeswaran2018AnEF}  & 60.0 & 43.4\\
\midrule
Our work  & \textbf{64.0} & \textbf{46.1} \\
\bottomrule

\end{tabular}
\par\end{centering}

\caption{The comparison with P@1 and P@5 between baseline systems from AskUbuntu.}
\label{tb:open_corpus}

\end{table}

\subsection{Main Results}

\subsubsection{Effect on AliExpress and LAZADA}

As shown in Table \ref{tb:top30}, we explore the effectiveness of our proposed approach on \textproc{Top-30} queries from Ali Express and LAZADA corpus. The presented architecture consists of original \textproc{CL} and \textproc{Auto-MLM}, and in this experiment we compare our model with highly analogous models and approaches used on the same task.
Due to the size of sessions and chats of Zh, En and Ar greater than other languages, almost all the baselines achieve remarkable results. 
By comparing \textproc{XLMR}$_{base}$ and \textproc{XLMR}$_{tiny}$, we find that the increase of the number of parameters could improve the performance of the model.
However, \textproc{XLMR}$_{base}$ shows less improvement than other models which are trained on sentence-level tasks.  
we obtain consistent better improvements on multi-lingual service corpora than any other methods.

\subsubsection{The Robustness}
To further verify the effectiveness of our model, we also validate the performance of our model on \textproc{Top-20} and \textproc{Top-10} queries from Ali Express and LAZADA corpus (see Figure \ref{lb_top_20_10}). 
On \textproc{Top-20} and \textproc{Top-10}, both the contrastive learning and the proposed method obtain highly similar even identical results on Ar.
On average, there are more questions (similar questions) for each knowledge configuration in Arabic,
and the coverage of query is relatively large. Thus, the performance of each model on Arabic is relatively good. 

\subsubsection{Verification on different corpus}
As shown in Table \ref{tb:open_corpus}, to further validate the performance of our method, we also make some explorations on openly available corpus AskUbuntu \citep{Lei2016SemisupervisedQR}. In this experiment we compare our approach with other recently presented highly similar  baselines. 
However, for fair comparison, we ignore the \textproc{XLMR}$_{base}$ model,  because there are big differences between the parameters of other baselines and \textproc{XLMR}$_{base}$.
Intuitively, our proposed method also shows better result than other SOTA methods on AskUbuntu with the evaluation metrics p@1 and p@5, i.e. one or five queries retrieved for the original user query.

\begin{figure*}[htp]
\centering 

\subfigure[]{
\begin{minipage}{7.1cm}
\centering

\includegraphics[width=7cm,height=3cm]{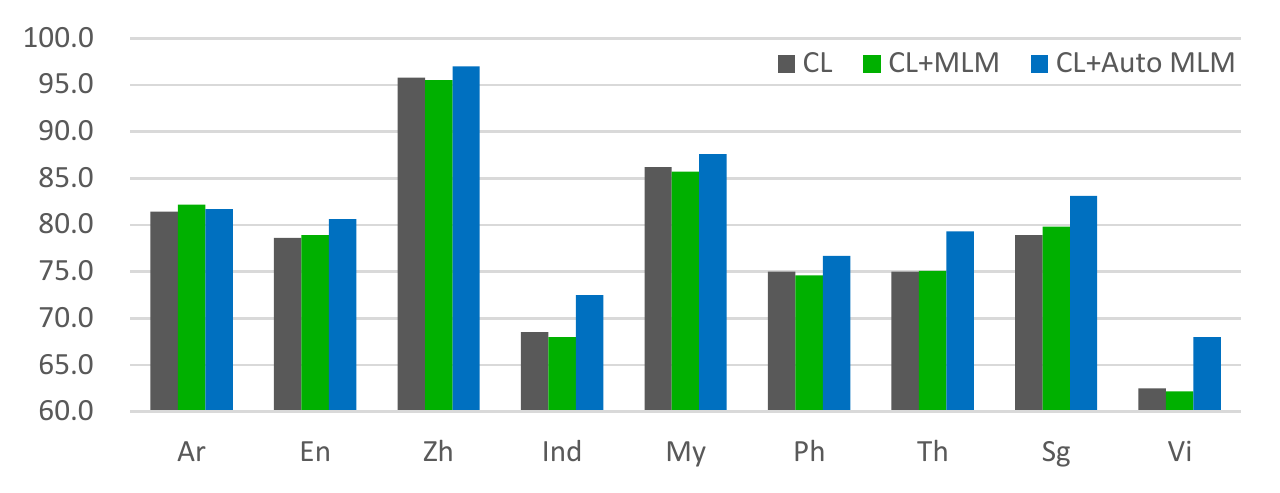}
\end{minipage}
}
\subfigure[]{
\begin{minipage}{7.1cm}
\centering
\includegraphics[width=7cm,height=3cm]{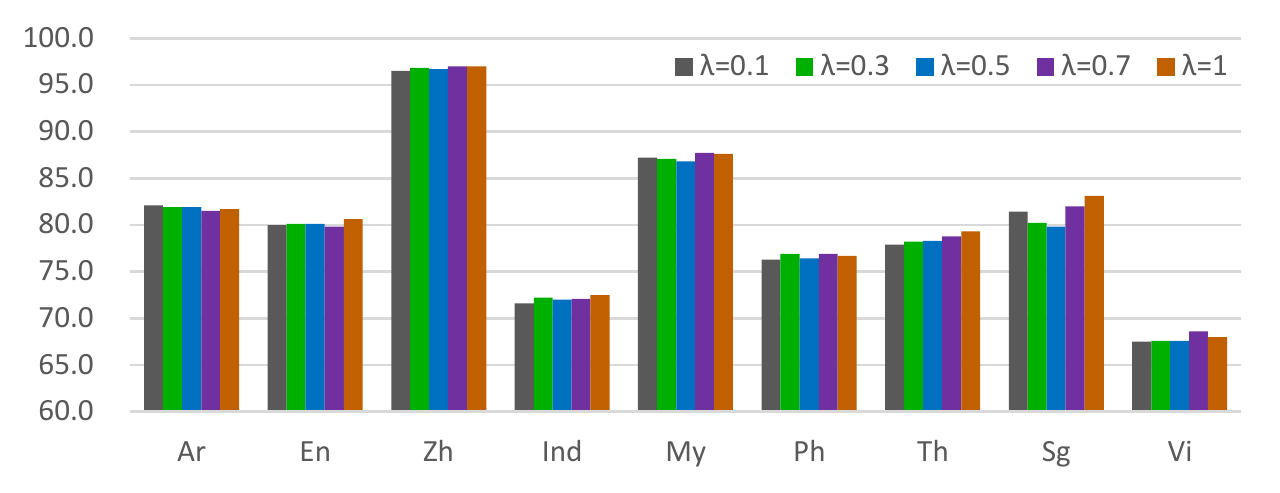}
\end{minipage}
}

\caption{(a) The comparison with and without \textproc{Auto-MLM} for our method on \textproc{Top-30} queries based on accuracy. (b) The effect of $\lambda$ to the performance of our proposed method on \textproc{Top-30} queries from Ali Express and LAZADA corpus based on accuracy (default value is $\lambda=1$).}

\label{fig:abl}
\end{figure*}

\begin{figure*}[!t]
\centering 

\subfigure[\textsc{\textproc{CMLM}}]{
\begin{minipage}{4.6cm}
\centering

\includegraphics[width=4cm,height=3.7cm]{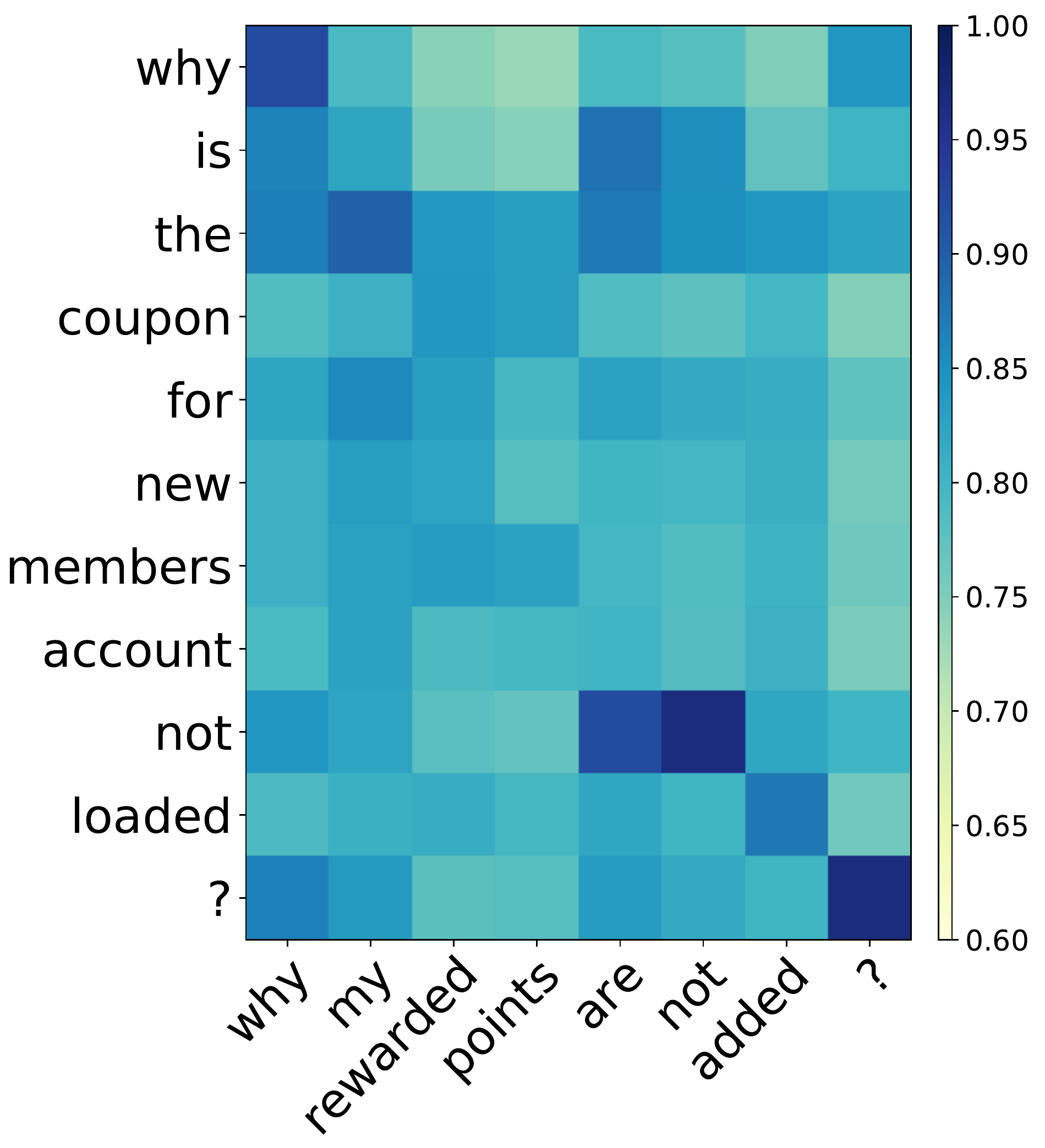}
\end{minipage}
}
\subfigure[\textsc{\textproc{CL}}]{
\begin{minipage}{4.6cm}
\centering
\includegraphics[width=4cm,height=3.7cm]{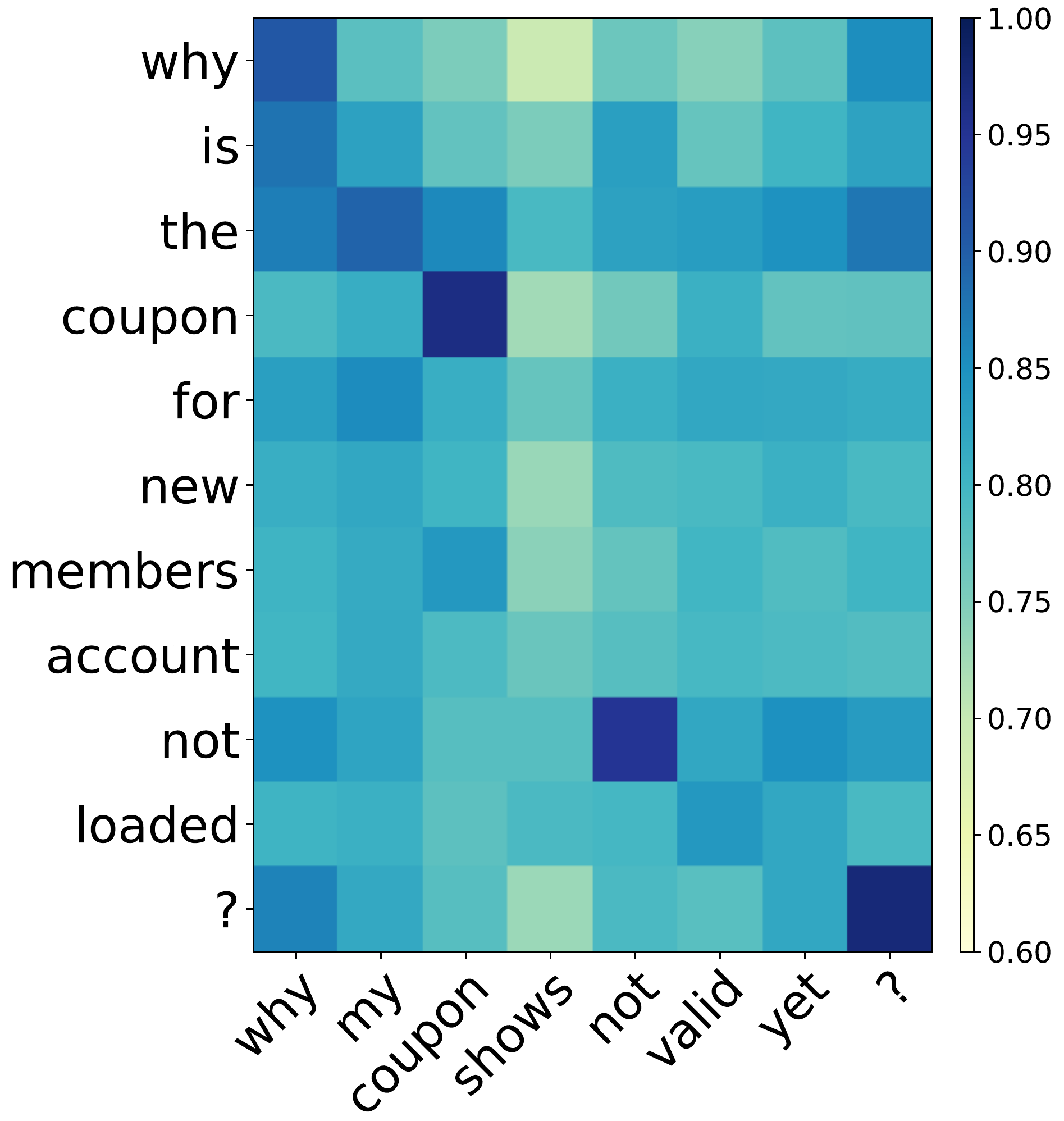}
\end{minipage}
}
\subfigure[\textsc{Our work}]{
\begin{minipage}{4.6cm}
\centering

\includegraphics[width=4cm,height=3.7cm]{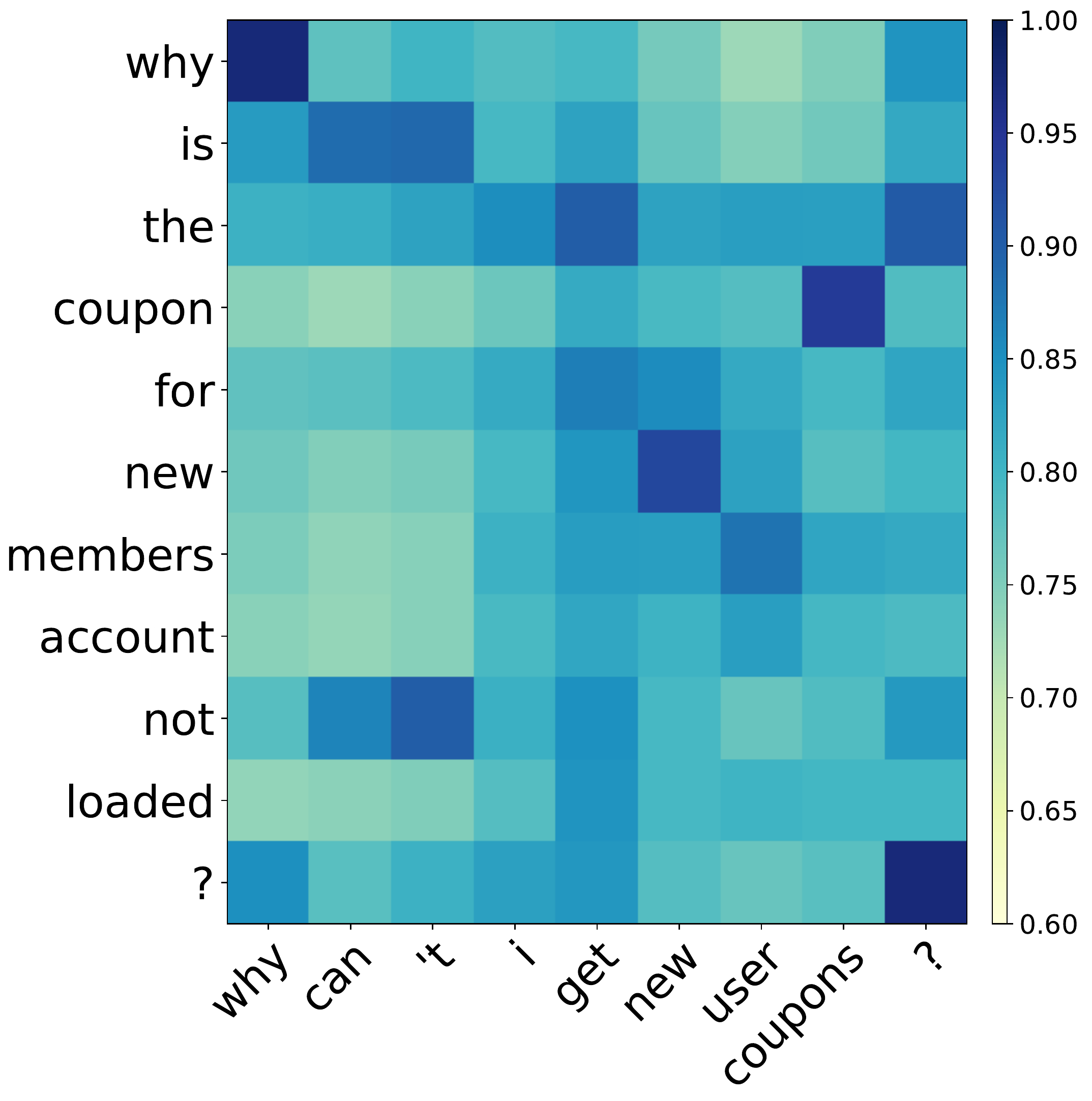}
\end{minipage}
}

\caption{The comparison of the retrieved similar samples between the strongest baselines and our work for user query, where ``y-axis" and ``x-axis" denote user query and retrieved knowledge from different systems.}

\label{lb_casestudy}
\end{figure*}

\subsubsection{Ablation Study}
\paragraph{Effect of with and without Auto-MLM} As depicted in Figure \ref{fig:abl}(a), we explore the effectiveness of with and without \textproc{Auto-MLM} for our model. Clearly, the proposed method consists of \textproc{CL} and \textproc{Auto-MLM}. Therefore, it is vital to investigate the effectiveness with and without \textproc{CL} and \textproc{Auto-MLM}.
Obviously, the combination of original \textproc{CL}  and \textproc{Auto-MLM} gains the consistent improvement on both corpora with average score. 
We also investigate the CL-MLM joint training method and observe that it does not show remarkable improvement. Our method outperforms both the CL and CL-MLM, which demonstrates that AutoMLM is an effective joint training approach.

\paragraph{Effect of $\lambda$ for improved Contrastive Learning} 
As given in methodology section , we jointly train our model by combining the contrastive loss and auto-mlm loss.
The parameter $\lambda$ plays an essential role, which can explain the importance of Auto-MLM (see Eq.\ref{eq_joint_loss}).
Intuitively, in our experiment we also make some investigations on this parameter and observe its effectiveness on the performance of proposed model. 
As illustrated in Figure \ref{fig:abl}(b), when the $\lambda = 1 $, our model achieves better performance on these corpora with the 0.81 accuracy.
It proves that Auto-MLM can effectively improve the QA retrieval results.

\subsubsection{Case Study}
Figure \ref{lb_casestudy} indicates the semantic similarities of retrieved knowledge samples. However, we only compare the better baselines CMLM and CL with our approach. 
It is effortless to find the baseline methods suffer from the semantic errors like  ``rewarded points'' and ``shows not valid'', but do not retrieve the knowledge which is same as the key idea of user query. 
The baseline CMLM also retrieves the samples with an incorrect subject ``rewarded points'' and also drops the subject (``new members'') of the user query. 
In addition, the baseline CL produces the sample  with semantically mismatched phrase ``shows not valid" at the end of the retrieved samples, while our method can extract the fine-grained coupon type ``new members" and retrieve the correct answer.

\section{Related Work}

\paragraph{CL}

The idea of contrastive learning can be traced back to 2006. \citet{DBLP:conf/cvpr/HadsellCL06} use CL to learn low-dimensional representations of high-dimensional data. Thereafter, CL has been also applied to several NLP tasks like word representation learning \citep{Mikolov2013EfficientEO} and knowledge graph representation learning \citep{DBLP:conf/nips/BordesUGWY13}. \citet{DBLP:conf/icml/ChenK0H20} prove that CL can achieve competitive performance compared to supervised learning on the ImageNet ILSVRC-2012 classification task.
CL has also received wide attention in the research field of NLP.
For example, \citet{Gao2021SimCSESC} use CL to learn sentence representations. It is also utilized in various NLP tasks like machine translation \citep{DBLP:conf/acl/PanWWL20} and dialog generation \citep{DBLP:conf/emnlp/CaiCSDBYZ20}.

\paragraph{MLM}

To overcome the major disadvantages of previous pretrained language models like ELMo \citep{DBLP:conf/acl/PetersABP17} and GPT \citep{GPT},
\citet{bert} propose a new training objective named masked language model (MLM) and utilize it in BERT. 
Then, multilingual MLMs like
\textproc{XLM-RoBERTa} \citep{Conneau2020UnsupervisedCR} are developed to handle NLP tasks in different languages. Thereafter, various studies have also proposed similar pretrained models like XLNet \citep{DBLP:conf/nips/YangDYCSL19} and BART \citep{DBLP:conf/acl/LewisLGGMLSZ20}.

\section{Conclusion and Future Work}

In this work we propose \textproc{Auto-MLM} approach to deal with the problem, which is overlooked by both contrastive learning and MLM. Precisely, MLM neglects the sentence level training and contrastive learning also ignores the extraction of internal info from the query.
In the future, we will try to employ the proposed approach on other NLP downstream tasks, such as representation learning, question answering, dialog generation, and dialog state tracking.

\bibliography{acl_latex}
\bibliographystyle{acl_natbib}

\end{document}